\documentclass[manuscript,screen]{acmart}
\AtBeginDocument{%
  \providecommand\BibTeX{{%
    \normalfont B\kern-0.5em{\scshape i\kern-0.25em b}\kern-0.8em\TeX}}}

\setcopyright{acmcopyright}
\copyrightyear{2023}
\acmYear{2023}
\acmConference[SARs: TMI at CHI 2023]{}{April 28, 2023}{Hamburg, Germany} 
\acmBooktitle{1st Workshop on Socially Assistive Robots as Decision Makers: 
Transparency, Motivations, and Intentions at CHI 2023, April 23--28, 2023, Hamburg, Germany} \acmPrice{15.00} \acmISBN{978-1-4503-XXXX-X/18/06}

\begin{document}

\title{Use of a Socially Assistive Robot as a Online Shopping Digital Skills Assistant}
\author{Scott MacLeod}
\email{sam19@hw.ac.uk}
\orcid{}
\affiliation{%
  \institution{Heriot-Watt University}
  \city{Edinburgh}
  \country{Scotland}
  \postcode{EH14 4AS}
}\author{Mauro Dragone}
\email{m.dragone@hw.ac.uk}
\orcid{}
\affiliation{%
  \institution{Heriot-Watt University}
  \city{Edinburgh}
  \country{Scotland}
  \postcode{EH14 4AS}
}


\begin{abstract}
      This work proposes and analyses the application of a robotic platform as an digital skills assistant. analysing the ethical issues relating to the decision making process in the use case of online food shopping in order to inform a co design session on what, and how, the digital skills assistant should make decisions.
\end{abstract}

\ccsdesc[300]{Human-centered computing~Accessibility technologies}
\ccsdesc[300]{Human-centered computing~Participatory design}
\ccsdesc[500]{Computer systems organization~Robotics}
\ccsdesc[500]{Human-centered computing~Personal digital assistants}
\ccsdesc[300]{Computer systems organization~Robotic autonomy}
\keywords{Socially assistive robots, Transparency, Robotic autonomy, Digital skills training}


\received{23 February 2023}

\maketitle

\section{Introduction}


Technology can assist older adults in a number of ways to enable healthy ageing, living independently, and to improve quality of life \cite{neves2022digital}.

Tele-medicine has recently gained popularity in healthcare \cite{neves2022digital}. Digital sessions reduce preparation and travel time for patients and has been found to improve patient outcomes and reduce hospital visits \cite{breen2015patient}.

Other digital solutions now allow treatments to occur in the patients home. Remote physio therapy has been tested where patients meet with a physio therapist, via tele-conference, to view live demonstratations ofcorrect exercise techniques, and then access a database of exercises with reference images and videos to follow while performing rehabilitation exercises without remote human supervision \cite{rasulnia2018assessing}.

The digital skills divide among older adults can limit their access to health services, as well as other services that rely on digital technology \cite{EUdigitaldivide2017}. This has the potential to exclude them from the ever-increasing number of services and advancements such as those mentioned above.

Digital technology has been found to be empowering for elderly users when they are designed with the elderly in mind \cite{blavzivc2020overcoming}, specifically by facilitating daily activities \cite{hill2015older}. Many physical challenging life factors, such as reduced mobility and social contact, can be addressed through the use of digital technology, motivating many programs aimed at increasing digital literacy in the elderly population such as \cite{martinez2021effects}.

\section{Background}

Digital skills training programs have been designed for individual training, peer tutoring, and class training in adult education scenarios. These methods have been applied in person, using a blended learning environment (BLE), and digitally. Objective measurements of digital literacy, such as the Digital Literacy Evaluation (DILE), have shown benefits from these approaches \cite{martinez2021effects}. Benefits of digital skills training has also been observed in terms of motivation and independence \cite{Pihlainen2021}. 

Many older adults might have obstacles such as health conditions and other difficulties that may prevent them from participating in digital skills training sessions. This motivated research into remote or automated digital skills training using a variety of technological methods, such as phone calls with human facilitators \cite{casselden2022not}, or conversational agents \cite{sriwisathiyakun2022enhancing} and using humanoid robots \cite{ijerph191710988}. 

In these scenarios the agents may operate either as assistants \cite{sriwisathiyakun2022enhancing} , i.e. helping people to carry out actual digital tasks, or as teachers \cite{ijerph191710988}, teaching people how to carry out the same tasks, by themselves. 


In \cite{ijerph191710988} a robot tutoring application was designed to give feedback through voice, facial expression, and gesture. Although the decision making for the robot was remotely controlled in this work, in a Wizard of Oz (WoZ) manner, they found that older adults thought that the robot tutor could be successfully used for different types of ICT training. They also found that voice and gestures increased satisfaction, and the perceived effectiveness of the tutoring, indicating for this type of application that a socially assistive robot (SAR) capable of this type of interaction would be preferable to that of a conversational agent. 

In these types of automated digital skills training scenarios the agents act as the facilitator for the digital skills training, providing prompts, suggestions, and choosing examples to perform as exercises in the training. In traditional digital skills training procedures, such as in \cite{martinez2021effects}, there are a large number of different scenarios considered, covering many aspects such as social media, virtual meetings, and online shopping. In all of these scenarios the facilitator would define the examples and walk though them with the participant. If these are to be facilitated using a SAR agent then many decisions would have to be made autonomously.

\section{Decision making of an online shopping assistant}

In the typical online shopping digital skills training the human facilitator would walk through an example purchase of one or many items from a shopping website. This involves showing all of the steps, such as logging in, finding the items, removing items from the basket, and reaching the final purchasing screen.

This scenario presents a number of different decisions that would need to be made by the facilitator that may have ethical implications. Some key decisions have been laid out in the following considering their implementation using a SAR:

\begin{itemize}
    \item \emph{What sites should be used for the training?} The decision on the specific stores that are used for training may influence older adults to use the same site in future purchases. If users are allowed to pick among different sites - something that can be easily facilitated in human-led digital skills training - then the SAR would need to have the capability to adapt to different layouts and workflows. These aspects may not be captured perfectly, if we consider that websites may be also modified and updated. This means that the SAR should also communicate how confident it is in the suggestions or assistance given, even acknowledging that it may make mistakes.

    In some cases the agent will need to communicate that it is having trouble identifying the assistance to give and to suggest trying a different items or different sites. The agent may also in this case proceed with a generic explanation of the step that it is expecting. For instance if it is unable to identify the location of a button or feature, but it expects it to be there with a specific name it may instruct the user to attempt to find it as well ('\emph{do you see a button?}'), as a form or cooperative peer training.

    \item \emph{What items should be used in the training example?} In the shopping scenario the agent will need to decide on a number of example recommendations for the participant to purchase, what examples does it chose and why, the examples that would traditionally be used would be relevant to the specific participants to the training session, in order to make the session more meaningful and to encounter and solve potential trip falls in those relevant purchases. The aim of determining items of specific relevant interest to the user is already done on shopping sites using recommendation algorithms. 
    
    The potential use of these recommendations to inform the example items included in a digital skills training session simplifies the task of the facilitator. However this may not be acceptable by the user group as an example as it is influenced by the recommendations of the site, which may have motivations to increase spending etc. In the case of a SAR assistant then it may not be obvious that this recommendation is done by the site not by the agent. If the agent is using external recommendations for purchases then at the very least that needs to be communicated to the participants in a clear way.
    
    \item If the facilitator is not using any external recommendations for the items that it is purchasing then this creates not only a technical challenge but a potential for bias towards certain perceived good choices. An example of this is making suggestions towards healthy eating during online food shopping.

\end{itemize}


We posit that in all of these cases then SAR's decisions should be bound within the preferences of the user group, as to avoid any situations where the robot makes a decision in a way that is unacceptable or unknown by the user. Within the ethical decision making in online shopping scenario, we propose the use of a co-design session to identify the preferences of the user group with respect to the above identified set of decisions, and to test the acceptability and usability of our application.

\section{Proposed Co-Design Study}

A co-design, or participatory design, workshop is a session with stakeholders, here older adults in digital skills training, to consult with, and aid in the design for the digital skills assistant.

The proposed co-design session is being designed in collaboration with a care provider\cite{blackwood} to inform the development of a prototype to be evaluated with a pilot user study.

In our first co-design session, we aim to define the preferences of older adults receiving digital skills training, on the types of decisions that a SAR facilitator in online shopping should make, and how it should communicate those decisions. 

 The user group will be recruited through the care provider and will include older adults who have already participated in one of their digital skills training sessions. The session will begin with demonstrations of different SAR platforms and their capabilities. This step is to familiarise the participants with different possible embodiment's for the SAR to be used for this application. 

Then participants will begin the discussion session where they will be shown examples of the decisions. Each decision will be proposed as a set of two examples, one for each of the different methods of decision making that has been proposed. In addition, the participants will be asked to put forward any additional methods of decision making that they would find preferable to the defined ones.

\begin{itemize}
    \item To identify different shops and apps that are used for online shopping by the user group, but also for the design of the desired assistance from the SAR. 

    \item To identify the method of picking example products to run through, and assist with the procedure to purchase those items, whether the user group would prefer pre-defined items, to ask during the session for particular items, or to use site recommendations.

    \item Whether in the assistance the user group would be interested in the facilitator suggesting items, either as the examples, or as alternatives to user defined examples. In addition to identifying suggestions by some objective preference, like price, or by other recommendation methods.

    \item To identify the preferred SAR embodiment and their interaction methods, whether to use voice, gestures, and expressions, or other modalities like assistive touch feature on tablet devices to highlight sections of the screen. 

\end{itemize}

After the first session the different outcomes will be collated as well and the participants encouraged to reflect and then send any thoughts or more comments, photo or note, email etc. These preferences will then inform the design of the application and this will be evaluated in a small pilot test with the same user group.

\section{Technical Architecture for Pilot test}

To conduct a pilot test of the decision making of the online shopping assistant then a technical architecture needs to be implemented. The proposed architecture for the training scenario is shown below in Figure \ref{fig:Architecture}. The SAR and the participant communicate using voice. The SAR can see how the user is interacting with the website by using a proxy, to intercept server-client interaction, and by controlling the users' tablet accessibility features, such as switch control, for instance to highlight or press buttons on the website. The proposed pilot study of the SAR will be conducted in a common room, providing digital skills assistance with a variety of participants.

\begin{figure}[h]
  \centering
  \includegraphics[width=0.5\linewidth]{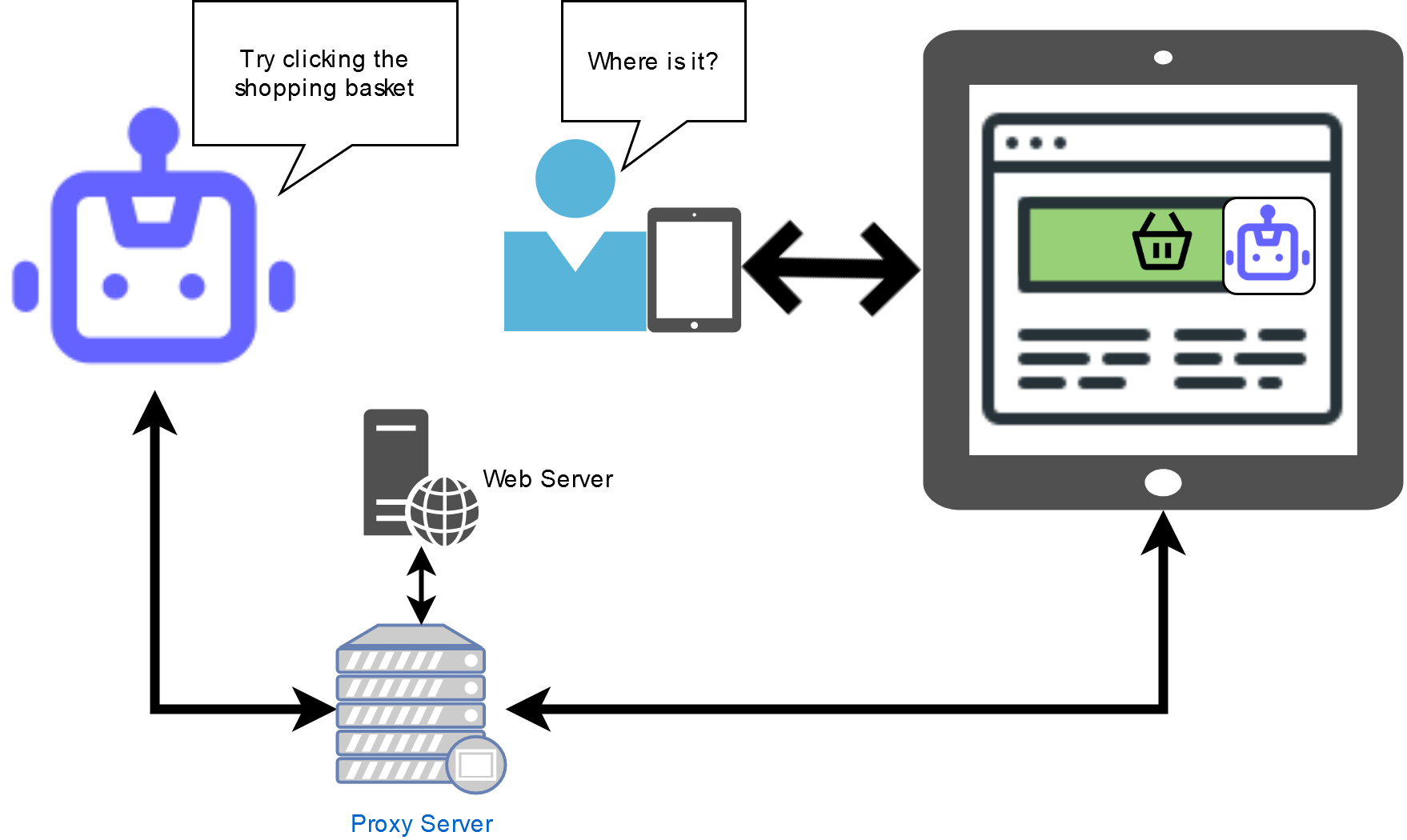}
  \caption{Proposed architecture of the digital skills training scenario showing the interaction of the SAR with the user, the application server and the client tablet.}
  \label{fig:Architecture}
  \Description{A robot asks the participant if they can click on the website's shopping basket. The participant asks where it is, and the robot highlights the basket on the tablet screen. The tablet and the robot are connected to share data bidirectionally through a proxy server, which is then connected to the web server.}
\end{figure}

Evaluation of this study will be measured with acceptance, usability, subjective performance, objective error data, and use data. We aim to test if analysing the potential decision making process with the relevant user group during the design of the system will result in a more acceptable and usable user experience.

\begin{acks}
This work was supported by funding from Innovate UK through the Knowledge Transfer Partnership (KTP) reference: 13068.
\end{acks}

\bibliographystyle{ACM-Reference-Format}
\bibliography{sample-base}


\begin{thebibliography}{12}


\ifx \showCODEN    \undefined \def \showCODEN     #1{\unskip}     \fi
\ifx \showDOI      \undefined \def \showDOI       #1{#1}\fi
\ifx \showISBNx    \undefined \def \showISBNx     #1{\unskip}     \fi
\ifx \showISBNxiii \undefined \def \showISBNxiii  #1{\unskip}     \fi
\ifx \showISSN     \undefined \def \showISSN      #1{\unskip}     \fi
\ifx \showLCCN     \undefined \def \showLCCN      #1{\unskip}     \fi
\ifx \shownote     \undefined \def \shownote      #1{#1}          \fi
\ifx \showarticletitle \undefined \def \showarticletitle #1{#1}   \fi
\ifx \showURL      \undefined \def \showURL       {\relax}        \fi
\providecommand\bibfield[2]{#2}
\providecommand\bibinfo[2]{#2}
\providecommand\natexlab[1]{#1}
\providecommand\showeprint[2][]{arXiv:#2}

\bibitem[Blackwood(2023)]%
        {blackwood}
\bibfield{author}{\bibinfo{person}{Blackwood}.}
  \bibinfo{year}{2023}\natexlab{}.
\newblock \bibinfo{title}{Blackwood Homes and Care}.
\newblock
\newblock
\urldef\tempurl%
\url{https://www.blackwoodgroup.org.uk/}
\showURL{%
\tempurl}
\newblock
\shownote{Accessed: 02-23}.


\bibitem[Bla{\v{z}}i{\v{c}} and Bla{\v{z}}i{\v{c}}(2020)]%
        {blavzivc2020overcoming}
\bibfield{author}{\bibinfo{person}{Borka~Jerman Bla{\v{z}}i{\v{c}}} {and}
  \bibinfo{person}{Andrej~Jerman Bla{\v{z}}i{\v{c}}}.}
  \bibinfo{year}{2020}\natexlab{}.
\newblock \showarticletitle{Overcoming the digital divide with a modern
  approach to learning digital skills for the elderly adults}.
\newblock \bibinfo{journal}{\emph{Education and Information Technologies}}
  \bibinfo{volume}{25} (\bibinfo{year}{2020}), \bibinfo{pages}{259--279}.
\newblock


\bibitem[Breen et~al\mbox{.}(2015)]%
        {breen2015patient}
\bibfield{author}{\bibinfo{person}{Sibilah Breen}, \bibinfo{person}{David
  Ritchie}, \bibinfo{person}{Penelope Schofield}, \bibinfo{person}{Ya-seng
  Hsueh}, \bibinfo{person}{Karla Gough}, \bibinfo{person}{Nick Santamaria},
  \bibinfo{person}{Rose Kamateros}, \bibinfo{person}{Roma Maguire},
  \bibinfo{person}{Nora Kearney}, {and} \bibinfo{person}{Sanchia Aranda}.}
  \bibinfo{year}{2015}\natexlab{}.
\newblock \showarticletitle{The Patient Remote Intervention and Symptom
  Management System (PRISMS)--a Telehealth-mediated intervention enabling
  real-time monitoring of chemotherapy side-effects in patients with
  haematological malignancies: study protocol for a randomised controlled
  trial}.
\newblock \bibinfo{journal}{\emph{Trials}} \bibinfo{volume}{16},
  \bibinfo{number}{1} (\bibinfo{year}{2015}), \bibinfo{pages}{1--17}.
\newblock


\bibitem[Casselden(2022)]%
        {casselden2022not}
\bibfield{author}{\bibinfo{person}{Biddy Casselden}.}
  \bibinfo{year}{2022}\natexlab{}.
\newblock \showarticletitle{Not like riding a bike: How public libraries
  facilitate older people’s digital inclusion during the Covid-19 pandemic}.
\newblock \bibinfo{journal}{\emph{Journal of Librarianship and Information
  Science}} (\bibinfo{year}{2022}), \bibinfo{pages}{09610006221101898}.
\newblock


\bibitem[Eurostat(2017)]%
        {EUdigitaldivide2017}
\bibfield{author}{\bibinfo{person}{Eurostat}.} \bibinfo{year}{2017}\natexlab{}.
\newblock \bibinfo{title}{A look at the lives of the elderly in the EU today}.
\newblock
\newblock
\urldef\tempurl%
\url{https://ec.europa.eu/eurostat/cache/infographs/elderly/index.html}
\showURL{%
\tempurl}
\newblock
\shownote{accessed: 02-23}.


\bibitem[Hill et~al\mbox{.}(2015)]%
        {hill2015older}
\bibfield{author}{\bibinfo{person}{Rowena Hill}, \bibinfo{person}{Lucy~R
  Betts}, {and} \bibinfo{person}{Sarah~E Gardner}.}
  \bibinfo{year}{2015}\natexlab{}.
\newblock \showarticletitle{Older adults’ experiences and perceptions of
  digital technology:(Dis) empowerment, wellbeing, and inclusion}.
\newblock \bibinfo{journal}{\emph{Computers in Human Behavior}}
  \bibinfo{volume}{48} (\bibinfo{year}{2015}), \bibinfo{pages}{415--423}.
\newblock


\bibitem[Jung et~al\mbox{.}(2022)]%
        {ijerph191710988}
\bibfield{author}{\bibinfo{person}{Sungwook Jung}, \bibinfo{person}{Sung~Hee
  Ahn}, \bibinfo{person}{Jiwoong Ha}, {and} \bibinfo{person}{Sangwoo Bahn}.}
  \bibinfo{year}{2022}\natexlab{}.
\newblock \showarticletitle{A Study on the Effectiveness of IT Application
  Education for Older Adults by Interaction Method of Humanoid Robots}.
\newblock \bibinfo{journal}{\emph{International Journal of Environmental
  Research and Public Health}} \bibinfo{volume}{19}, \bibinfo{number}{17}
  (\bibinfo{year}{2022}).
\newblock
\showISSN{1660-4601}
\urldef\tempurl%
\url{https://doi.org/10.3390/ijerph191710988}
\showDOI{\tempurl}


\bibitem[Mart{\'\i}nez-Alcala et~al\mbox{.}(2021)]%
        {martinez2021effects}
\bibfield{author}{\bibinfo{person}{Claudia~I Mart{\'\i}nez-Alcala},
  \bibinfo{person}{Alejandra Rosales-Lagarde}, \bibinfo{person}{Yonal~M
  P{\'e}rez-P{\'e}rez}, \bibinfo{person}{Jose~S Lopez-Noguerola},
  \bibinfo{person}{Mar{\'\i}a~L Bautista-D{\'\i}az}, {and}
  \bibinfo{person}{Raul~A Agis-Juarez}.} \bibinfo{year}{2021}\natexlab{}.
\newblock \showarticletitle{The effects of Covid-19 on the digital literacy of
  the elderly: norms for digital inclusion}. In
  \bibinfo{booktitle}{\emph{Frontiers in Education}}, Vol.~\bibinfo{volume}{6}.
  Frontiers Media SA, \bibinfo{pages}{716025}.
\newblock


\bibitem[Neves et~al\mbox{.}(2022)]%
        {neves2022digital}
\bibfield{author}{\bibinfo{person}{Ana~Lu{\'\i}sa Neves},
  \bibinfo{person}{Charilaos Lygidakis}, \bibinfo{person}{Kyle Hoedebecke},
  \bibinfo{person}{Lu{\'\i}s de Pinho-Costa}, {and} \bibinfo{person}{Alberto
  Pilotto}.} \bibinfo{year}{2022}\natexlab{}.
\newblock \showarticletitle{Digital health in an ageing world}.
\newblock \bibinfo{journal}{\emph{The Role of Family Physicians in Older People
  Care}} (\bibinfo{year}{2022}), \bibinfo{pages}{107--118}.
\newblock


\bibitem[Pihlainen et~al\mbox{.}(2021)]%
        {Pihlainen2021}
\bibfield{author}{\bibinfo{person}{Kaisa Pihlainen}, \bibinfo{person}{Kristiina
  Korjonen-Kuusipuro}, {and} \bibinfo{person}{Eija Kärnä}.}
  \bibinfo{year}{2021}\natexlab{}.
\newblock \showarticletitle{Perceived benefits from non-formal digital training
  sessions in later life: views of older adult learners, peer tutors, and
  teachers}.
\newblock \bibinfo{journal}{\emph{International Journal of Lifelong Education}}
  \bibinfo{volume}{40}, \bibinfo{number}{2} (\bibinfo{year}{2021}),
  \bibinfo{pages}{155--169}.
\newblock
\urldef\tempurl%
\url{https://doi.org/10.1080/02601370.2021.1919768}
\showDOI{\tempurl}
\showeprint{https://doi.org/10.1080/02601370.2021.1919768}


\bibitem[Rasulnia et~al\mbox{.}(2018)]%
        {rasulnia2018assessing}
\bibfield{author}{\bibinfo{person}{Mazi Rasulnia},
  \bibinfo{person}{Billy~Stephen Burton}, \bibinfo{person}{Robert~P Ginter},
  \bibinfo{person}{Tracy~Y Wang}, \bibinfo{person}{Roy~Alton Pleasants},
  \bibinfo{person}{Cynthia~L Green}, {and} \bibinfo{person}{Njira Lugogo}.}
  \bibinfo{year}{2018}\natexlab{}.
\newblock \showarticletitle{Assessing the impact of a remote digital coaching
  engagement program on patient-reported outcomes in asthma}.
\newblock \bibinfo{journal}{\emph{Journal of Asthma}} \bibinfo{volume}{55},
  \bibinfo{number}{7} (\bibinfo{year}{2018}), \bibinfo{pages}{795--800}.
\newblock


\bibitem[Sriwisathiyakun and Dhamanitayakul(2022)]%
        {sriwisathiyakun2022enhancing}
\bibfield{author}{\bibinfo{person}{Kanyarat Sriwisathiyakun} {and}
  \bibinfo{person}{Chawaporn Dhamanitayakul}.} \bibinfo{year}{2022}\natexlab{}.
\newblock \showarticletitle{Enhancing digital literacy with an intelligent
  conversational agent for senior citizens in Thailand}.
\newblock \bibinfo{journal}{\emph{Education and Information Technologies}}
  \bibinfo{volume}{27}, \bibinfo{number}{5} (\bibinfo{year}{2022}),
  \bibinfo{pages}{6251--6271}.
\newblock


\end{thebibliography}

\end{document}